\documentclass[conference]{IEEEtran}
\IEEEoverridecommandlockouts

\usepackage{cite}
\usepackage{amsmath,amssymb,amsfonts}
\usepackage{algorithmic}
\usepackage{graphicx}
\usepackage{textcomp}
\usepackage{xcolor}
\usepackage{multirow}
\usepackage{hhline}
\usepackage{url}
\def\BibTeX{{\rm B\kern-.05em{\sc i\kern-.025em b}\kern-.08em
    T\kern-.1667em\lower.7ex\hbox{E}\kern-.125emX}}
\begin{document}
\renewcommand{\arraystretch}{1.25}

\title{Target word activity detector: An approach to obtain ASR word boundaries without lexicon\\
}

\author{ Sunit Sivasankaran, Eric Sun, Jinyu Li, Yan Huang, Jing Pan

\IEEEauthorblockN{}
\IEEEauthorblockA{
\textit{Microsoft Speech and Language Group}\\
One Microsoft Way, Redmond, WA, USA \\
\{sunit.sivasankaran,ersun,jinyli,yanhuang,jingpan\}@microsoft.com}
}

\maketitle

\begin{abstract}
Obtaining word timestamp information from end-to-end (E2E) ASR models remains challenging due to the lack of explicit time alignment during training. This issue is further complicated in multilingual models. Existing methods, either rely on lexicons or introduce additional tokens, leading to scalability issues and increased computational costs. In this work, we propose a new approach to estimate word boundaries without relying on lexicons. Our method leverages word embeddings from sub-word token units and a pretrained ASR model, requiring only word alignment information during training. Our proposed method can scale-up to any number of languages without incurring any additional cost. We validate our approach using a multilingual ASR model trained on five languages and demonstrate its effectiveness against a strong baseline.

\end{abstract}


\section{Introduction}

Downstream Automatic speech recognition (ASR) tasks, such as speaker diarization \cite{park2022review} and voice editing, frequently depend on precise word timestamp information. However, this information is challenging to obtain from end-to-end (E2E) ASR models \cite{li2022recent, prabhavalkar2023end}, including Connectionist Temporal Classification (CTC) models \cite{graves2006connectionist,Miao2015EESENES,ctc_li}, attention-based encoder-decoder (AED) models \cite{cho2014learning, Attention-bahdanau2014, chan2016listen}, and transducer models like recurrent neural network transducers (RNN-T) \cite{Graves2012SequenceTW, he2019streaming, Li2020Developing} and transformer-transducers (T-T) \cite{zhang2020transformer, chen2021developing}. These models typically lack explicit time alignment information during training, which impedes their ability to reliably predict the start and end times of the recognized words. This challenge is exacerbated in multilingual models that are designed to transcribe multiple languages using a single model.

To address this issue, various methods have been proposed in the literature \cite{9576572,chen21j_interspeech,7952667,9054250,sainath20_interspeech, zhao2021addressing}. For example, Zhao et al. \cite{zhao2021addressing} suggested estimating the probability of context-independent (CI) phones per acoustic frame. This approach utilized a simple linear layer to project an ASR encoder embedding onto the CI phone space. During evaluation, the phonetic expansion of the word sequence was generated, and a Viterbi algorithm was used to align the estimated CI phones with the phone sequence derived from the lexicon. This straightforward method proved surprisingly effective, providing reliable word boundary estimates at a low computational cost. However, its dependence on the lexicon made it challenging to scale to multilingual models. Additionally, this method in the multilingual setup requires language identification information for each word to select the appropriate lexicon, which is not always reliable.

Alternative approaches to word time estimation that do not rely on a lexicon have also been explored. Sainath et al. \cite{sainath20_interspeech} applied constraints during training to enhance word timing accuracy. They introduced an extra token, ``word boundary" to mark the word start time and used the last word piece to indicate the word end time. While this method gave reliable word timing accuracy, it increased training costs and might cause significant ASR accuracy degradation due to the added ``word boundary" token. The SubWord Alignment Network (SWAN) proposed by Kang et al. \cite{Kang2024} is an interesting approach aimed to estimate word timing without phonetic information. However, it requires alignment at the subword-unit or token level, which is harder to obtain when scaling the model across multiple languages. The model also relies on  a voice activity detector (VAD) to manage silence frames, which can further increase computational costs. Jiang et al.\cite{jiang23e_interspeech} proposed directly predicting the duration of each word using an aligner module.  A novel cost function was proposed where the model directly predicts the normalized duration of the word instead of predicting the presence of a word or sub-unit. The aligner model was however trained only on a monolingual ASR model.

In this work, we draw inspiration from the speaker diarization community to propose a method for obtaining word timing without relying on lexicons. The target speaker voice activity detector (TS-VAD) \cite{tsvad_medennikov,wang_tsvad} uses speaker embeddings and Mel filterbanks as inputs to determine speaker activity. In our method, we instead learn word embeddings using sub-word token units and a pretrained ASR model as part of our proposed `Target word activity detector' (TWAD) model. Using these embeddings we estimate `word activities', or the acoustic frames at which the word is active, and derive word boundaries in an audio segment.

Our approach can scale to any number of languages without the need for any type of lexicons. The TWAD model only requires word alignment information during training, which is easier to obtain compared to the phonetic or sub-word level alignment needed in previous methods. To validate our approach, we tested it using a multilingual ASR model trained on five languages: English (EN), French (FR), Spanish (ES), Italian (IT), and German (DE). We also compare the word time estimation metrics against a strong baseline.

\section{TWAD model architecture}
\begin{figure*}
    \centering
    \includegraphics[width=0.9\linewidth]{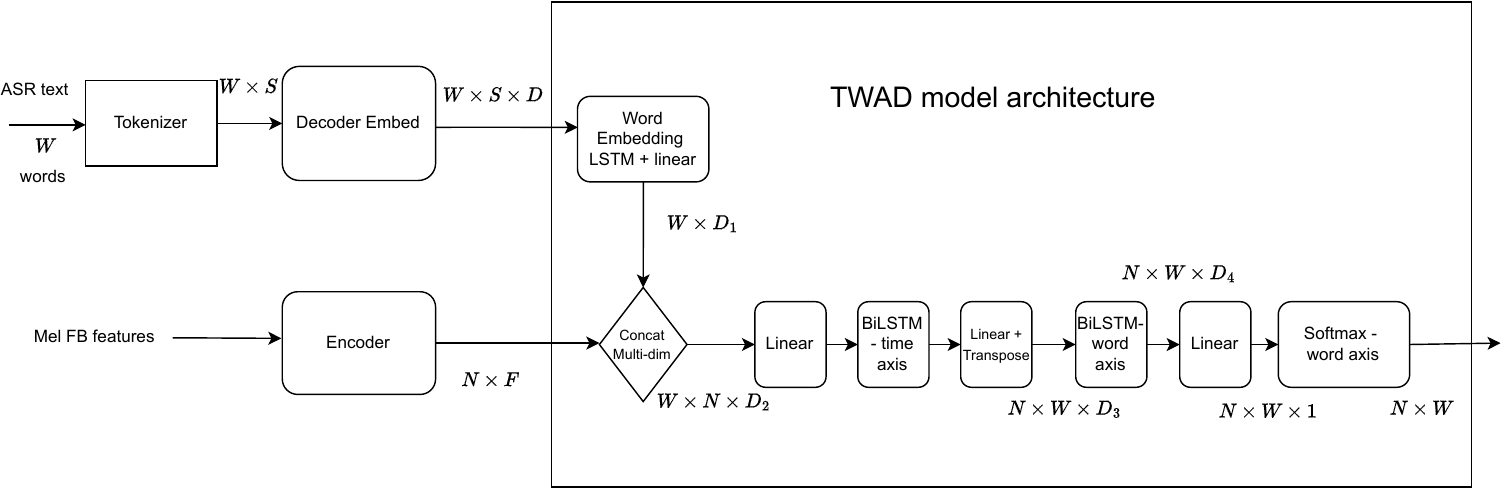}
    \caption{TWAD model architecture. Only parameters inside the TWAD model are updated. $W$ is the number of words in a sentence. $S$  is the max number of tokens across all the words in a sentence.}
    \label{fig:ph_free_model}
\end{figure*}

The Target Word Activity Detector (TWAD) model is designed to identify word activity within a given speech signal. The model architecture is in Fig. \ref{fig:ph_free_model}.  The input to the model is the encoder embedding, which can be derived from any layer of the encoder. Let $\mathbf{x}$ be the output of the encoder embedding for the speech signal:
$
    \mathbf{X} = \{\mathbf{x}_1, \hdots, \mathbf{x}_n, \hdots, \mathbf{x}_N\} \in \mathbb{R}^{N\times F}
$, where $N$ represents the total number of frames and $F$ denotes the dimension of the encoder embedding.

The Automatic Speech Recognition (ASR) transcription, denoted as $\mathbf{t}$, contains $W$ words:
$    
    \mathbf{t} = \{w_1, \hdots, w_i, \hdots,  w_W\}.
$
Each word $w_i$ in the sentence $\mathbf{t}$ can be decomposed into a set of tokens using a tokenizer. If $\mathbf{w}_i$ is the vector representing the tokens of the word $w_i$, then
    $\mathbf{w}_i = \{s_{i_1}, \hdots, s_{i_S}\}$, 
where $S$ is the maximum number of tokens. Consequently, the entire sentence $\mathbf{t}$ can be represented by a set of tokens:
$ 
    \mathbf{t} = \{ \{s_{1_1}, \hdots, s_{1_S}\}, \hdots, \{s_{W_1}, \hdots, s_{W_S}\}\} $.

The decoder embedding is used to generate a sentence representative matrix $\mathbf{T} \in \mathbb{R}^{W\times S \times D}$, where $D$ is the token embedding dimension. The TWAD model takes $\mathbf{X}$ and $\mathbf{T}$ as input features and estimates a word activity matrix $\hat{\mathbf{A}} \in \mathbb{R}^{N \times W}$.

The first module of the TWAD model is the word embedding estimator, which employs a bidirectional Long Short-Term Memory (BiLSTM) network. This module estimates an embedding for every word in the sentence using the token embeddings from the matrix $\mathbf{T}$. The final output from the bidirectional LSTM is concatenated and passed through a linear layer, resulting in an embedding for each word in the sentence, effectively collapsing the token dimension in the process. Importantly, the model does not require explicit token alignment information during training, as the objective is to derive word embeddings based on the tokens. Moreover, acquiring precise token level alignment data is inherently challenging.

The word embeddings are concatenated with each frame of the encoder embedding $\mathbf{X}$ to create a combined 3-dimensional representation that includes both word and acoustic embeddings. This process involves replicating each $\mathbf{x}_n$ by a factor of $W$, resulting in a matrix of shape $\mathbb{R}^{W \times N \times F}$. The word embedding is then concatenated to form a joint representation matrix with the shape $\mathbb{R}^{W \times N \times (F + D_1)}$. Finally, this matrix is projected onto a smaller dimensional space $D_2$ using a linear layer. To capture the temporal correlation between features, a BiLSTM is applied along the time axis of the 3-dimensional matrix, followed by another BiLSTM along the word axis to model correlations between words. A linear layer is applied to reduce the final embedding to a single dimension. A softmax activation function applied along the word dimension, yields the probability of each word, $p(w_i|\mathbf{x}_n)$.

Cross-entopy is used as loss function to train the TWAD model, which is defined as:
\[
    L(\mathbf{A}, \hat{\mathbf{A}}) = \sum^N_{n=1} \sum^W_{i=1} \mathbb{I}_n(w_i) \log(p(w_i|\mathbf{x}_n)
\]
where $\mathbf{A}$ is a matrix of size $\mathbb{R}^{N \times W}$ containing the ground truth word activity information, and $\mathbb{I}_n(w_i)$ is an indicator variable that signals the presence of word $w_i$ in the $n$-th frame.

\begin{figure}
    \centering
    \includegraphics[width=0.8\linewidth]{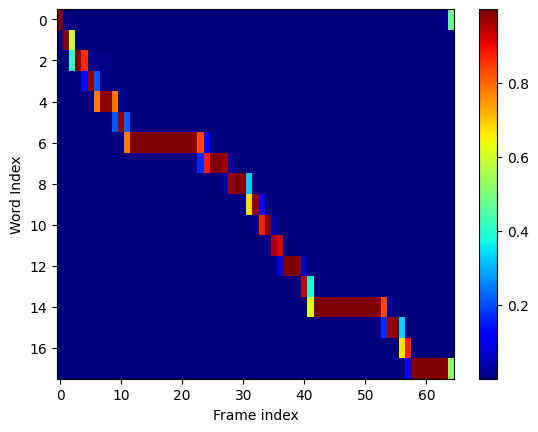}
    \caption{Output of the TWAD model for the sentence:  `Je vais te poser une question, je veux que tu me dises la vérité, tu es d'accord?' which after text normalization becomes   `je vais te poser une question je veux que tu me dises la vérité tu es d'accord'}
    \label{fig:word_activity}
\end{figure}
The TWAD model can be trained to estimate any number of word activities in an audio segment, but for efficient batch creation, the maximum value of $W$ is set to 100. The first word in $\mathbf{A}$ is always set to silence. The rest of the words in $\mathbf{A}$ are arranged in the same sequence as they appear in the sentence. An example of the estimated $\mathbf{\hat{A}}$ matrix is as shown in Fig. \ref{fig:word_activity}. 

The estimated word timing matrix $\mathbf{\hat{A}}$, can be noisy. To  obtain the word boundaries from $\mathbf{\hat{A}}$, a discrete time warping (DTW) algorithm is used. An optional silence is inserted between every word before DTW. This allows for pauses or silence in-between words or towards the beginning or end of the utterance. 


\section{Experimental settings}

\subsection{ASR Model }

The ASR model is an audio encoder (AED) \cite{chan2016listen,li20_interspeech} consisting of nemo-convolution layers followed by 24 conformer \cite{conformer_gulati} layers. The nemo-convolution layers \cite{nemo_conv,nemo_doc} takes in 80-dimension Mel-filterbanks computed using a shift of $10$ms and window  of $25$ms as input and sub-sample the time frame by a factor of 8 in order to reduce the computation cost. This implies that the word change detection can happen at a granularity of only $80$ ms. Each conformer layer has a multi-head attention with 16 heads, and a depth-wise convolution with kernel size of 3. The multi-head attention and the depth-wise convolution are sandwiched between two 1536-dim feed-forward layers. The decoder consists of 3 conformer layers and the feed-forward layer dimension is 4096. The embedding dimension is 1024. The AED is trained to optimize the combined cross-entropy loss and the CTC loss (weighted by 0.2). 


We train the AED ASR  model on 5 different languages for modeling English, Italian, French, German and Spanish. The training data detail is described in \cite{xue2023weakly}. The transcripts preserve capitalization, punctuation without text normalization. We pool the transcripts of training data and train a byte pair encoding (BPE) tokenizer with a vocabulary size of 15911. In order to reduce the language confusion and improve training stability, the model is trained to predict the language IDs in the output transcript as a prefix token. We add the language IDs as special tokens in the vocabulary list such as  $<en>$, $<it>$, $<fr>$, $<de>$, $<es>$.

\subsection{Baseline word time estimation model}

The baseline model, described in \cite{zhao2021addressing}, calculates the probability of context-independent phones at each frame. This probability is estimated by a straightforward linear layer that uses the 9th layer embedding of the frozen ASR encoder as input. Layer 9 was determined to produce the most effective results compared to other layers tested.

The training of the baseline CI phone prediction model utilized a dataset comprising five languages: English (EN), French (FR), Spanish (ES), Italian (IT), and German (DE). The number of training hours per language was as follows: 75K for EN, 25K for FR, 28K for ES, 41K for IT, and 12K for DE. Across these languages, there are a total of 420 context-independent (CI) phones, with 129 for EN, 99 for FR, 30 for ES, 110 for IT, and 52 for DE.  The training data underwent text normalization, followed by the application of a force alignment model to determine phonetic and word boundaries.

During the evaluation of the context-independent phone word estimation model, referred to as the baseline in this work, the estimated language identification (LID) of the word was used to select the phonetic expansion based on the corresponding language lexicon. The discrete time warping (DTW) algorithm was employed to align the expected phone sequence with the estimated phone sequence. Optional silence was inserted between words.

\subsection{TWAD model}
\label{subsec:twad}

The BiLSTM used to obtain the word embedding consists of 512 hidden units. The final output from each BiLSTM layer is concatenated and projected to a 512-dimensional space through a linear layer. For modeling temporal correlations, two BiLSTM layers, each with 512 hidden units, are applied along the time axis. Additionally, a single BiLSTM layer with 64 hidden units is used to capture word correlations within a sentence. The final linear layer projects the 64-dimensional embedding to a 1-dimensional scalar value. A dropout rate of 0.2 is consistently applied across all BiLSTM layers. The first word is always set to silence.  The same dataset used for the baseline CI phone based model was used to train the TWAD model.

\subsection{Evaluation}

The ASR model generates output in a display-format text designed for readability, often including punctuation and numbers as they appear in spoken language. For example, the text `two hundred dollars' would appear in display format as `$\$200$'. However, for both the baseline and TWAD model,  we preprocess this output. We transform the ASR-generated text by normalizing it, which involves removing or replacing non-textual elements such as punctuation and numbers with their textual equivalents. This ensures that the hypothesis text aligns with the text-normalized data used during training, making it suitable for evaluation and comparison.  

An alternative approach  particularly for the TWAD model, is to train using the display format text. The additional punctuation information contained in text can only improve the word timing model. This approach will be investigated in our future work. Note that neither the baseline model nor the TWAD model updates the parameters of the ASR model; thus, neither has an impact on the performance of the ASR system.

 Internal Microsoft datasets were used to evaluate both the word time estimation models.  The reference word boundary is obtained by force alignment using a traditional hybrid model. There are 5 different datasets, one each for the languages mentioned above. The dataset contains utterances across challenging acoustic conditions and speaking styles. Only the words which were correctly identified by the ASR model was used to compute the word timing metrics for both the models. A summary of number of hours, utterances, word counts and word error rate (WER) are shown for each language in Table \ref{tab:eval_dataset}.

 \begin{table}[]
 \centering
\caption{Summary of datasets used to evaluate word timing model across languages}
\begin{tabular}{|l|cccc|}
\hline
  Languages & \multicolumn{1}{l|}{\#Hours} & \multicolumn{1}{l|}{\#Utterances} & \multicolumn{1}{l|}{\#Words} & \multicolumn{1}{l|}{WER\%}\\ \hline \hline

EN & 5.4                      & 113                      & 23548  &      11.9              \\ \cline{1-1}
FR & 5.2                      & 6378                     & 45753    &    9.8              \\ \cline{1-1}
ES & 4.2                      & 8852                     & 23236   &    3.1               \\ \cline{1-1}
IT & 6.4                      & 6378                     & 29827    &    5.6              \\ \cline{1-1}
DE & 10.1                       & 4492                     & 53064   &   6.9                \\ \cline{1-1}
\hline
\end{tabular}
\label{tab:eval_dataset}
\end{table}

\subsection{Metric}
When estimating word boundaries, it's crucial to consider both the start and end times of the words. For these timestamps, we calculate the absolute differences between the estimated and ground truth word times and then determine the average, p50, p90, and p95 values across all the words of the evaluation set.

\section{Results and discussion}

\begin{table}
\centering
\caption{Start and end time word delta statistics in milliseconds.}
\scalebox{0.9}{
\begin{tabular}{|l|l|l|lllll|l|} 
\hline
\textbf{Model}                     & \textbf{Delta}                             & \textbf{Metric} & \multicolumn{1}{l|}{\textbf{EN}} & \multicolumn{1}{l|}{\textbf{FR}} & \multicolumn{1}{l|}{\textbf{ES}} & \multicolumn{1}{l|}{\textbf{IT}} & \multicolumn{1}{l|}{\textbf{DE}} & \textbf{Ave.} \\ 
\hhline{|=========|}

\multirow{8}{*}{\textbf{Baseline}} & \multirow{4}{*}{\textit{Start }} & \textit{Average}  & 40.6                             & 56.5                             & 43.5                             & 52.7                             & 45.4   &47.7      \\ 
\cline{3-3}
                                   &                                       & \textit{p50}      & 30                               & 40                               & 40                               & 30                               & 40        & 36   \\ 
\cline{3-3}
                                   &                                       & \textit{p90}      & 70                               & 90                               & 70                               & 80                               & 80       & 78    \\ 
\cline{3-3}
                                   &                                       & \textit{p95}      & 100                              & 140                              & 90                               & 150                              & 120     & 120     \\ 
\cline{2-9}

                                   & \multirow{4}{*}{\textit{End }}   & \textit{Average}  & 52.0                             & 52.4                             & 44.2                             & 55.6                             & 40.4     & 48.9    \\ 
\cline{3-3}
                                   &                                       & \textit{p50}      & 30                               & 50                               & 40                               & 30                               & 30     & 36      \\ 
\cline{3-3}
                                   &                                       & \textit{p90}      & 100                              & 100                              & 70                               & 90                               & 70      & 86     \\ 
\cline{3-3}
                                   &                                       & \textit{p95}      & 180                              & 120                              & 90                               & 180                              & 90       & 132    \\ 
\hline

\multirow{8}{*}{\textbf{TWAD}}     & \multirow{4}{*}{\textit{Start }} & \textit{Average}  &  61.6                               & 51.8                             & 35.7                             & 42.9                             & 34.9  & 45.3       \\ 
\cline{3-3}
                                   &                                       & \textit{p50}      &   40                               & 50                               & 30                               & 30                               & 30    & 36       \\ 
\cline{3-3}
                                   &                                       & \textit{p90}      &     90                             & 100                              & 60                               & 70                               & 60    & 76       \\ 
\cline{3-3}
                                   &                                       & \textit{p95}      &       110                           & 120                              & 70                               & 90                               & 80   & 94        \\ 
\cline{2-9}

                                   & \multirow{4}{*}{\textit{End }}   & \textit{Average}  &         90.6                         & 65.5                             & 44.0                             & 58.8                             & 34.5    & 58.7     \\ 
\cline{3-3}
                                   &                                       & \textit{p50}      &             30                     & 50                               & 30                               & 30                               & 30  & 34          \\ 
\cline{3-3}
                                   &                                       & \textit{p90}      &               110                   & 120                              & 70                               & 100                              & 60    & 92       \\ 
\cline{3-3}
                                   &                                       & \textit{p95}      &            290                      & 170                              & 120                              & 220                              & 80  & 176         \\
\hline
\end{tabular}
}
\label{tab:results}
\end{table}

\subsection*{Comparison with respect to baseline:}
The results obtained using both the baseline and TWAD model for all 5 languages are as shown in Table \ref{tab:results}. The average start and end time deltas for baseline model is around $50$ ms and the p95 values are in the acceptable range for downsteam tasks such as speaker diarization. This can be considered as a strong baseline. The start deltas of the TWAD model is better than the baseline model as evidenced by the average, p50 and p90  values across all languages. Similar trend can be observed for end delta values in most cases. 

It is important to note that apart from the word boundary metrics, the TWAD model outperforms the baseline model in several key aspects. Unlike the baseline model, the TWAD model does not require any lexical information, allowing it to scale to any number of languages without increasing computational complexity. In contrast, the baseline model depends on additional lexicon information to align the lexicon-generated phone sequence with the predicted phone sequence to obtain the word boundary. As the number of languages increases, the number of output CI phone dimensions also increases, leading to confusion between phones and a subsequent deterioration in word boundary estimation metrics. In spite of these constrains, the fact that the metrics of TWAD model are comparable to baseline model clearly shows the superiority of the proposed method.



\subsection*{Mismatched conditions:}

As mentioned in Section \ref{subsec:twad}, the ASR model was trained to output display text and the TWAD model was trained on tokens obtained from normalized text data. If we evaluate the TWAD model using the tokens generated from display text, we see a degradation in the word timing performance.  For example, the average start and end delta pair for DE and IT was $(49.2, 51.5)$  and $ (58.2, 74.3)$  ms  respectively using the token obtained from display format text compared to $(34.9, 34.5)$ and $ (42.9, 58.8)$ ms obtained after normalizing the ASR hypothesis.


\subsection*{Impact of different layer encoder embeddings as acoustic input: }

The start and end delta average for EN using the TWAD model was $(110.2, 121.0)$ ms, $(195.7,175.5)$ ms, and $(248.2,264.9)$ ms for layers $22$, $20$, and $9$, respectively. This performance is significantly worse than the $(61.6,90.6)$ ms achieved with the last layer (layer $23$), as shown in Table \ref{tab:results}. This observation can be explained by the increasing correlation between the encoder and decoder embeddings as we move up the layers.

In contrast, the CI phone baseline model yielded the best results when using the 9th layer of the encoder as acoustic input, with progressively worse results observed as we moved up or down the layers. This phenomenon is likely due to the loss of phoneme information in the higher layers, where the embedding is closer to the token embedding and not enough phonetic information in the lower layers.

\section{Conclusion}

We introduced a method to determine word boundaries without relying on subunit or phonetic level alignments. This approach has demonstrated scalability for multilingual models. Experiments conducted on English, French, Spanish, Italian, and German languages indicate that the word time estimation errors are comparable to those of models that utilize additional lexicon information, which our proposed model does not use. To enhance the model's performance, we plan to incorporate punctuation and related subword token units during training.

\section*{Acknowledgment}

Thanks to Amit Das and Rui Zhao for discussion regarding ASR decoder and phone alignment models.

\newpage
\bibliographystyle{IEEEbib}
\bibliography{refs}

\end{document}